\documentclass[letterpaper, 10 pt, conference]{ieeeconf}  

\IEEEoverridecommandlockouts                              

\usepackage{graphics} 
\usepackage{epsfig} 
\usepackage{mathptmx} 
\usepackage{times} 
\usepackage{amsmath} 
\usepackage{amssymb}  
\usepackage{balance}

\usepackage{hyperref}
\usepackage{algorithm}
\usepackage{algpseudocode}
\usepackage{multirow}
\usepackage[table]{xcolor}
\usepackage{pifont}
\newcommand{\cmark}{\ding{51}}%
\newcommand{\xmark}{\ding{55}}%

\title{\LARGE \bf
Low-Cost Teleoperation with Haptic Feedback through Vision-based Tactile Sensors for Rigid and Soft Object Manipulation
}

\author{Martina Lippi*$^{1}$, Michael C. Welle*$^{2}$, Maciej K. Wozniak$^{2}$, Andrea Gasparri$^{1}$, Danica Kragic$^{2}$
\thanks{*These authors contributed equally (listed in alphabetical order).}
\thanks{ ${}^1$Roma Tre University, Rome, Italy  {\it\small \{martina.lippi,andrea.gasparri\}@uniroma3.it} }%
\thanks{ ${}^2$KTH Royal Institute of Technology Stockholm, Sweden, {\it\small \{mwelle,maciejw,dani\}@kth.se}}%
}

\begin{document}

\maketitle
\thispagestyle{empty}
\pagestyle{empty}

\begin{abstract}
Haptic feedback is essential for humans to successfully perform complex and delicate manipulation tasks. A recent rise in tactile sensors has enabled robots to leverage the sense of touch and expand their capability drastically. However, many tasks still need human intervention/guidance. For this reason, 
we present a teleoperation framework designed to provide haptic feedback to human operators based on the data from camera-based tactile sensors mounted on the robot gripper. Partial autonomy is introduced to prevent slippage of grasped objects during task execution.
Notably, we rely exclusively on low-cost off-the-shelf hardware to realize an affordable solution.
We demonstrate the versatility of the framework on nine different objects ranging from rigid to soft and fragile ones, using three different operators on real hardware.

\end{abstract}


\section{Introduction}
Tactile sensors are designed to replicate the sense of touch in machines, enabling robots to gather information about the physical properties of objects they come in contact with, including texture, shape, and softness. This tactile capability is especially critical when executing precise and delicate manipulations of fragile objects, such as handling food items or brittle materials. Among the various tactile sensor technologies \cite{yamaguchi2019recent}, camera-based tactile sensors have recently gained considerable interest within the research community thanks to their high resolution, robustness, circuit simplicity, and low cost~\cite{vbsensors_review} compared to electronic ones. 
Such sensors generally comprise a deformable elastomer integrated with a camera positioned behind to capture high-resolution images of the deformation occurring during contact.  
To name a few applications, in-hand manipulation \cite{yamaguchi2019tactile}, volumetric mesh reconstruction \cite{zhu_icra2022}, and manipulation of soft and brittle objects \cite{welle2023enabling} have been accomplished in the literature using autonomous robots equipped with vision-based tactile sensors. However, in many real-world applications, human guidance remains indispensable to ensure successful outcomes. This necessity arises in scenarios such as recovering from unexpected failures, operating in highly unstructured environments, or recording example manipulations.

Motivated by the above observations, our study delves into a teleoperation scenario where we aim to combine human guidance skills with the tactile information gathered from vision-based tactile sensors mounted on the robot end effector. More specifically, we propose a novel  Tactile-to-Haptic (T2H) algorithm where we translate the observed tactile information to vibrations received on the user's teleoperation controller. 
Although haptic feedback has been largely applied in the literature, e.g., see \cite{teleop_survey_tro,el2020review,bolopion_TASE2013},
to the best of our knowledge, there exist no readily deployable solutions for bringing camera-based tactile sensor data into affordable consumer-grade hardware that can be used to teleoperate a robotic manipulator.
Our framework aims to fill this need by providing an easy-to-deploy way for teleoperating a Franka Panda manipulator, equipped with a gripper and two DIGIT tactile sensors \cite{lambeta2020digit}, using an Oculus Quest 2 controller. the gripper. 
The 6D velocities from the hand-held controller are mimicked by the robot end effector, while the trigger buttons are used to close and open the gripper. 
The tactile sensing is translated into the amplitude of the perceived vibration of the controller giving the user an intuitive feeling of if, and how strongly, an object is grasped. Furthermore, we equip the framework with a partial autonomy behavior where the robot automatically detects and prevents slippage once an object is being grasped.
\begin{figure}
    \centering
    \includegraphics[width=\linewidth]{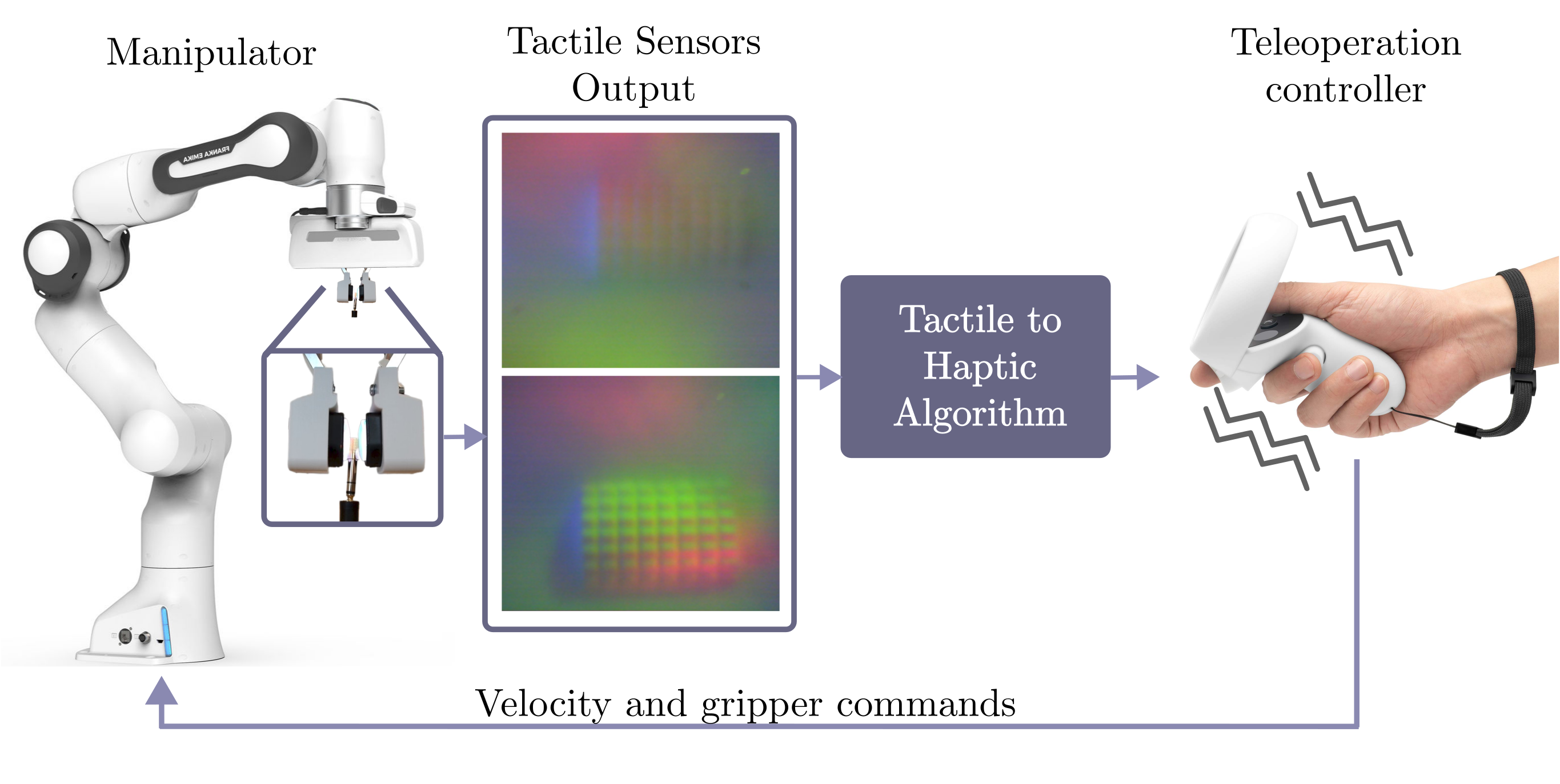}
    \caption{Overview of the proposed T2H teleoperation framework: the data obtained with camera-based tactile sensors are elaborated by the T2H algorithm, the resulting haptic feedback is provided to the operator via the teleoperation controller, and the commands for the robot are generated.}
    \label{fig:overview}
\end{figure}
In detail, our contributions are:
\begin{itemize}
    \item A novel T2H teleoperation framework deployable on low-cost hardware that translates tactile sensor readings into haptic vibration feedback, while providing partial autonomy for slippage prevention.
    \item Versatile demonstrations by three different operators for the use of such haptic feedback with and without partial autonomy for rigid and soft object manipulation.
    \item Public available source code\footnote{\fontsize{6}{8}\url{https://github.com/vision-tactile-manip/quest2_ros_msg}} and detailed setup instructions\footnote{\fontsize{6}{8}\url{https://vision-tactile-manip.github.io/teleop/setup-instructions}} for reproducibility purposes. 
\end{itemize}

\section{Related Work}
Remote robot control, often referred to as teleoperation is an active area of research and there are various approaches on how to do it most efficiently and accurately~\cite{darvish2023teleoperation}. Xie et al. discussed the use of joysticks as a control interface~\cite{xie2023design}. Ishiguro et al. delved into the application of exoskeletons \cite{ishiguro2020bilateral}, while Gliesche's work compared keyboard and mouse with kinesthetic devices~\cite{gliesche2020kinesthetic}. Nandikolla et al. showed that another approach to control the robots can be via brain-computer interfaces (BCI) as a means to command robots using users' brain signals, allowing them to interact with robots in an innovative way~\cite{nandikolla2022teleoperation}. 
While these approaches offer advanced teleoperation, it is important to note that they often involve expensive equipment, and require specialized knowledge and training. Additionally, most of them are heavily tailored to specific robotic platforms.

On the other hand, Virtual, Augmented, and Mixed (VAM) technologies offer advanced and immersive applications that can be used to manipulate robots via position or velocity controllers, 
and can be easily adapted to different robotic platforms~\cite{wozniak2023virtual,barentine2021vr, xu2022shared,moletta2023virtual}. Moreover, VAM frameworks have been introduced to facilitate human-robot collaboration by visualizing the states and intentions of robots in delivery tasks~\cite{chandan2021arroch,wozniak2023happily}.scenarios~\cite{ostanin2018interactive}.

However, to the best of our knowledge, none of these VAM frameworks focuses on manipulating soft and fragile objects using haptic feedback. 
Our method uses a VAM framework controller to provide the user with a sense of touch recorded via tactile sensors. 
This sensorial capability comes at a low hardware cost while enhancing the ease of manipulation for the user, especially when handling fragile objects. 

\subsection{Haptic feedback in human-robot interaction}

Haptic feedback plays an important role in achieving precision in robot manipulation. It encompasses various feedback options and controllers that contribute to improving the user's understanding of the robot's state and movement. Various feedbacks are classified as haptics. Vibrotactile feedbacks are vibrations that can provide sensory information about whether the robot is closer or further from an obstacle~\cite{lorenzini2022performance,chua2023modular} while squeezing bands can simulate the sensation of grasping objects~\cite{smith2022feeling}. 
Six DoF controllers apply resistance forces to guide the robot's movements, intervening, for instance,  if the robot deviates from the intended path~\cite{luzfeeling,pocius2020communicating}. However, those solutions are highly specialized and expensive. Additionally, the implementation of a \textit{guiding force} as haptic feedback can sometimes lead to confusion and conflicts between the user and the robot, and finding the balance is an active area of research~\cite{al2023resolving}.
Moreover, haptic devices include gloves equipped with resistance plates to simulate the sensation of holding objects~\cite{hinchet2018dextres}, wrist-worn haptic devices that offer controller-less haptic feedback for virtual and augmented reality experiences~\cite{palmer2022haptic}, and devices designed for simulating interactions with virtual objects, with haptic feedback concentrated on the wrist~\cite{coffey2022collaborative,stanley2012evaluation}.

While these applications focus on the haptic feedback to improve the precise manipulation, there are few works on using visual sensors (such as DIGIT~\cite{lambeta2020digit}, Minsight~\cite{andrussow2023minsight}, or GelSight~\cite{yuan2017gelsight}) as a driver for the haptic feedback. Cao et al.~\cite{cao2023vis2hap} use visual sensors employed in the robotic arm to assess material properties, surface characteristics, roughness, and friction, and translate it to the haptic screen that reflects the properties of this material.

In our paper, we take a different approach. First of all, our solution uses VAM controllers and visual sensors that are commercially available, affordable, and not robot-specific. Next, we use  these visual sensors to correlate the haptic feedback with how much force is applied to the object the robotic arm is grasping. This helps the operator to understand how much force is applied to the object, preventing unnecessary object compression that may cause damage to it, but also preventing \textit{slippage} when not enough force is applied. 

\section{Problem statement and method overview}

The primary objective of this study is to develop a teleoperation framework that leverages readily available and cost-effective hardware to manipulate objects that can possibly be delicate. Specifically, the framework should achieve the following goals: \emph{i) cost-effective teleoperation,} i.e., realizing a system based on low-cost and commercial hardware components, enhancing its affordability; \emph{ii) delicate object manipulation}, i.e., enabling the precise and gentle manipulation of fragile objects; 
\emph{iii) user-friendly interface}, i.e., designing an intuitive interface for users, ensuring ease of use, and reducing the learning curve for operators of varying expertise levels; \emph{iv) real-time control}, i.e., providing control and feedback mechanisms with no delay. 

To achieve the above objectives, we propose the T2H framework depicted in Fig.~\ref{fig:overview}. Specifically, we consider a robotic manipulator equipped with off-the-shelf vision-based tactile sensors on the gripper, and a controller able to provide vibration feedback to the user. 
In our setup, we employ the following commercial and low-cost devices: DIGIT tactile sensors \cite{lambeta2020digit} for the robot and an Oculus Quest 2 controller for teleoperation. 
The tactile sensor data are translated into vibration feedback based on low-computational image processing, allowing the user to perceive the sense of the applied pressure on the object under manipulation. Concurrently, the  controller's linear and angular velocities 
are transformed into velocity commands for the robot's end effector,  providing a reactive and intuitive interface for the users. In addition, we design a partial autonomy behavior to proactively prevent object slippage during manipulation. In practice, this entails the automatic tightening of the grasp whenever autonomous slippage is detected, ensuring a secure hold on the object. 
The synergy between the haptic feedback 
and the autonomous behavior empowers users with the ability to manipulate delicate objects with ease (as shown in Sec~\ref{sec:exp}),  reducing the risk of excessive compression of these objects. 
Note that the proposed framework can be adapted to any vision-based tactile sensor and teleoperation controller with vibration feedback.

\section{T2H Framework}
 
The proposed T2H framework comprises the following key modules which are detailed in the rest of the section: \emph{i)} mapping from tactile sensors to teleoperation controller, 
    \emph{ii)} mapping from teleoperation controller to robot commands, and \emph{iii)} 
partial autonomy for slippage prevention.

\begin{figure}
    \centering
    \includegraphics[width=\linewidth]{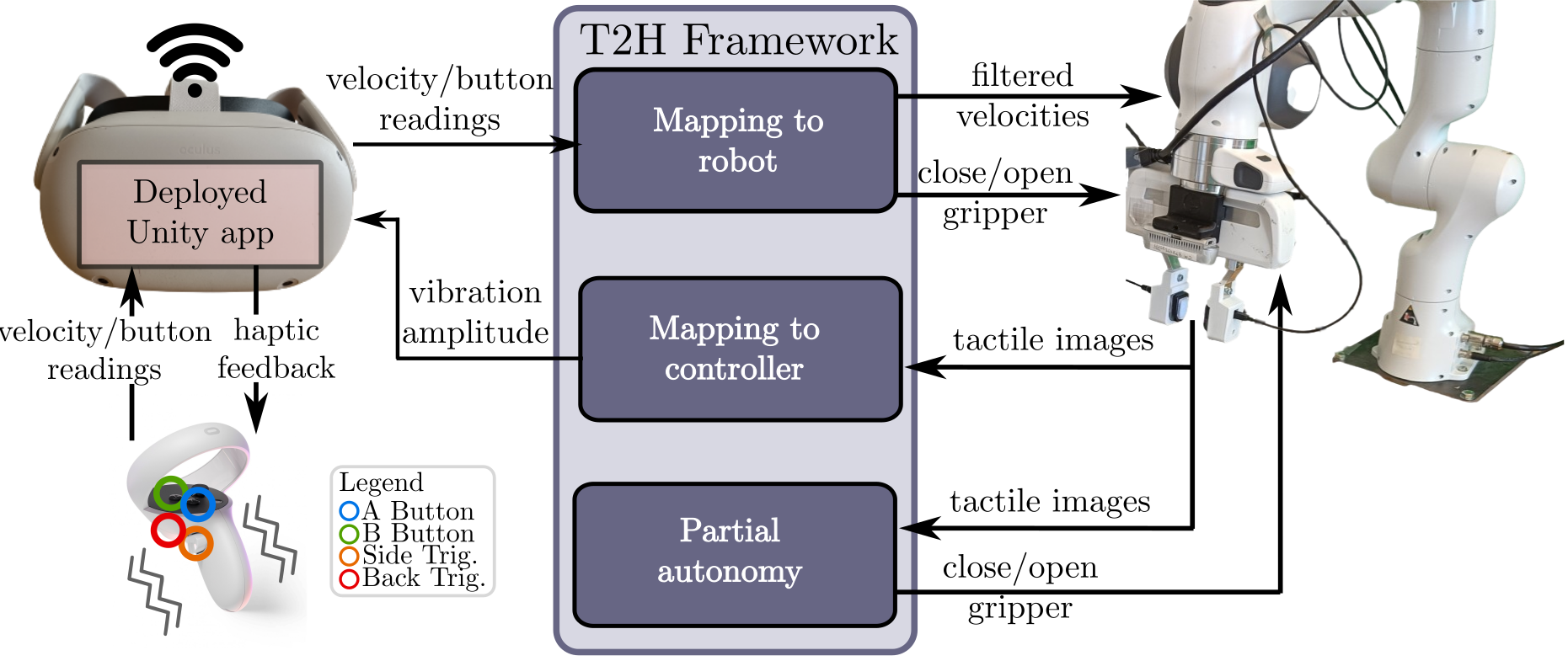}
    \vspace{-10pt}
    \caption{Communication architecture for the T2H teleoperation framework. 
    }
    \vspace{-11pt}
    \label{fig:arch}
\end{figure}

The communication architecture, along with the information flow, is illustrated in Fig.~\ref{fig:arch} and is fully integrated in ROS middleware. 
We  leverage the Unity Robotics toolbox\footnote{\fontsize{6}{8}\url{https://github.com/Unity-Technologies/Unity-Robotics-Hub}}  
 which enables communication between Unity and ROS. 
Specifically, it allows us to create a custom Unity application that is 
deployed on the Oculus headset and is capable of sending/receiving information to/from ROS. In this way, 
 haptic feedback can be provided to the teleoperation controller, while also allowing commands from the controller to be forwarded to the robot.
 In the proposed framework, the operator has a direct view of  the operating environment and is not required to wear the headset. However, the development of an additional graphical user interface (GUI) to further augment human perception is planned for future work.

\subsection{Mapping from tactile sensors to haptic feedback}

To determine the haptic feedback given the tactile sensors' images, we expand upon the processing methodology introduced in our prior work \cite{welle2023enabling}. 
The basic idea is to compute the pixel-wise variation for each sensor image with respect to a background reference image,   which can be set by the operator at any time, and then relate the vibration feedback with the count of altered pixels. In general, as the grasping force intensifies, the proportion of pixels differing from the background image also increases, resulting in amplified haptic feedback that should be conveyed to the operator.

\begin{figure}
    \centering
    \includegraphics[width=\linewidth]{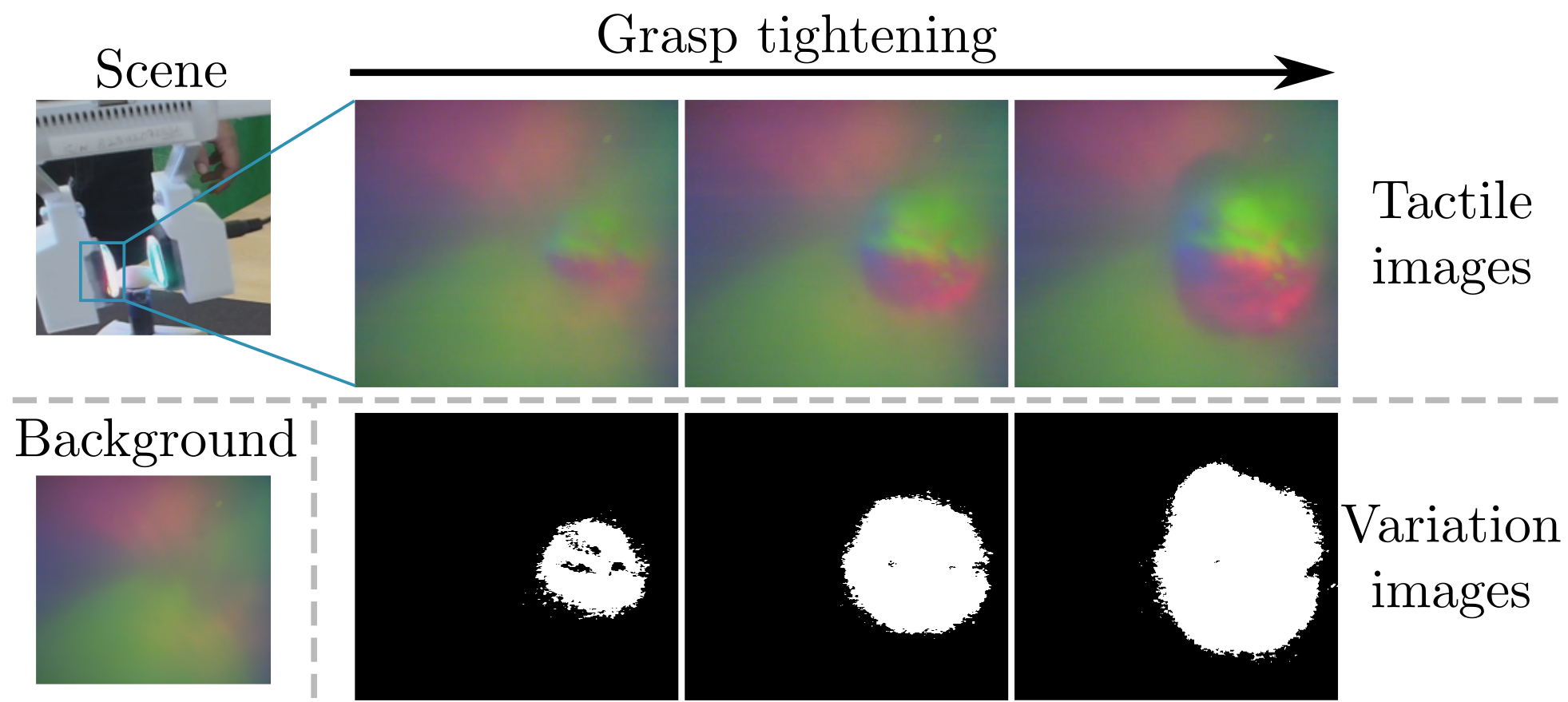}
    \vspace{-10pt}
    \caption{Representation of the tactile images (first row) with respective variation images (second row) as the grasp is tightened on a pistachio nut. The scene and the background are shown on the left.
    }
    \vspace{-11pt}
    \label{fig:diff}
\end{figure}

We refer to the pixel-wise variation in each sensor as variation image and compute it as reported in Algorithm~\ref{alg:tcd}. First, we evaluate the pixel-wise difference between the current sensor image and the background as $D^s_k = |T^s_k - R^s|, $ (line 1) where $T^s_k$ is the RGB image of the tactile sensor $s$, with $s\in\{1,2\}$, at time step $k$, $R^s$ is the RGB reference image of the sensor, and $D^s_t$ is the resulting RGB difference image. Then, the three RGB channels are averaged to produce a one-dimensional difference image \mbox{$\bar{D}^s_k$} (line 2), which undergoes a thresholding operation to reduce noise in the tactile sensor data. More precisely, we set pixel values to $0$ if their intensity falls below an automatically tuned noise threshold $\eta^s\in[0,1]$, and to $1$ otherwise. In this way, a binary difference image ${B}_{k}^s$ is generated (line 3) which captures the significant differences between the current tactile image and the reference image. Details about the computation of the background reference image and the tuning of threshold noise are provided later in the section. For a more reliable (but slower) touch detection process, the variation image $V^s_k$ is obtained with the element-wise (or Hadamard) product of $c$ consecutive binary difference images (lines~4-7),~i.e., 
$$
V^s_k= {B}_{k}^s \odot {B}_{k-1}^s ... \odot {B}_{k-c+1}^s, 
$$
with $\odot$ denoting the Hadamard product, i.e., to classify a pixel as a variation from the reference image, it must exhibit this variation for a minimum of $c$ consecutive frames.
The percentage of variation pixels, i.e., equal to $1$, is finally computed and normalized in the range $[0,1]$ (line 8). 
Figure~\ref{fig:diff} shows the tactile images (first row) and the respective variation images (second row) obtained as the grasp is tightened on a pistachio nut. The background is reported on the left. 

\begin{figure}
    \centering
    \includegraphics[width=\linewidth]{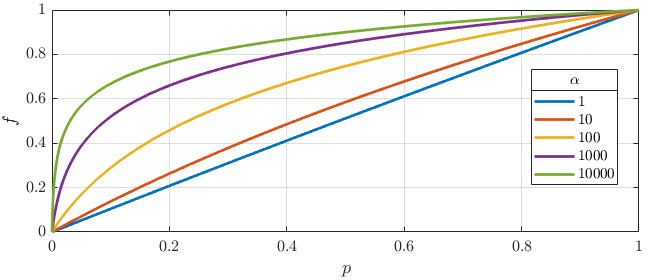}
    \vspace{-22pt}
    \caption{Representation of the haptic feedback with different gains  $\alpha$.}
    \label{fig:log}
    \vspace{-11pt}
\end{figure}

\begin{algorithm}
\caption{Tactile variation detection}\label{alg:tcd}
    \begin{algorithmic}[1]
    \Require $T^s_k, [B^s_{k-c+1},..., B^s_{k-1}], R^s, \eta^s$
    \State $D^s_k = |T^s_k - R^s|$
    \State $\bar{D}^s_k =$average channels($D^s_k$)
    \State $B^s_k =$ thresholding($\bar{D}^s_k$, $\eta^s$)
    \State $V^s_k = B^s_k$
    \For{$i \in \{k-c+1, ..., k\}$}:
        \State $V^s_k = V^s_k \odot B^s_i$
    \EndFor
    \State $p^s_k=$percentage variation pixel($V^s_k$)/100
    \State\Return $p^s_k,B^s_k$    
    \end{algorithmic}
\end{algorithm}
Based on the above, 
our T2H algorithm, reported in Algorithm 
\ref{alg:t2h}, generates the vibration feedback that is provided to the user. More specifically, at each time step $k$, the tactile images are retrieved (line 3), and the variation ratios $p^s_k$ 
are computed, with $s=1,2$ (lines 4-5).
At this point, the haptic feedback $f_k$, representing the controller vibration intensity, is determined  with the following logarithmic function (line~7)
$$f_k = \log_{10}(1 + \alpha p_k) / \log_{10}(1 + \alpha), $$
where $p_k$ is the average variation ratio (line 6) and $\alpha$ is a positive constant playing a critical role in the feedback behavior as shown in  Fig.~\ref{fig:log}: 
as $\alpha$ increases, the haptic feedback becomes more sensitive to small changes in $p_k$, leading to a steeper incline for low values of $p_k$. Conversely, as $\alpha$ decreases, 
a flatter slope for low values of $p_k$ is obtained. Finally, the haptic feedback is sent to the teleoperation controller (line 8) and the lists of previous binary difference images are updated (lines 9-10). 

\begin{algorithm}
    \caption{Tactile to Haptic (T2H) mapping}\label{alg:t2h}
    \begin{algorithmic}[1]
        \Require $R^s, \eta^s$ with $s=1,2$, $\alpha,c$
        \State $L^1=\emptyset, \,L^2=\emptyset$
        \For{time step k}
        \State $T^1_k,T^2_k =$collect tactile images
            \State $p^1_k, B^1_k=$tactile variation detection($T^1_k,L^1,R^1,\eta^1$)
        \State $p^2_k, B^2_k=$tactile variation detection($T^2_k,L^2,R^2,\eta^2$)
        \State $p_k=(p^1_k+p^2_k)/2$
    \State $f_k = \log_{10}(1 + \alpha p_k) / \log_{10}(1 + \alpha)$
    \State send haptic feedback($f_k$)    
    \State $L^1=$add image to fixed-size list$(B^1_k,c)$
    \State $L^2=$add image to fixed-size list$(B^2_k,c)$
        \EndFor      
    \end{algorithmic}
\end{algorithm}

To establish the reference images  $R^s$ and  automatically set the noise thresholds $\eta^s$
with $s=1,2$, we ask the operator to press the \textit{B} button (marked in green in Fig.~\ref{fig:arch})  before starting the manipulation (i.e., in no contact condition). This preliminary step triggers the collection of  $K$  consecutive tactile images for each sensor: the simple average of the first $N$ images forms the reference image, while 
the average maximum value across the one-dimensional difference images \mbox{$\bar{D}^s_k$} establishes the noise threshold, 
 i.e., $\eta^s=\frac{1}{K}\sum_{i=k}^{k+K} \max(\bar{D}^s_i). $

\subsection{Mapping from teleoperation controller to robot}
The purpose of the module for mapping from the teleoperation controller to the robot is to replicate with the robot the movements executed by the operator. This translation of commands persists as long as the \textit{A} button remains pressed, i.e., the robot receives zero-velocity commands in the case the button is not pressed.
When the button \textit{A} is pressed, both the linear and rotational velocities of the teleoperation controller are mapped into inputs for the low-level Cartesian controller of the manipulator. 
For safety reasons, a sliding window averaging filter with length $w$ is additionally applied to smooth thewe velocity inputs, and each component of the linear velocity is bounded within the range $[- v^l_{max}, v^l_{max}]$, with $ v^l_{max}$ a positive constant. Similarly, the angular velocity components are restricted to the range $[- v^r_{max}, v^r_{max}]$, with $ v^r_{max}$ a positive constant. Moreover, velocity components within the range of $[- v^l_{min}, v^l_{min}]$ for linear motion and $[- v^r_{min}, v^r_{min}]$  for rotational motion, with $v^l_{min}$ and $v^r_{min}$ positive constants,  are filtered out and are set to zero. This ensures that the controller does not respond to minor velocity changes. 
To command the gripper, the back trigger (in red in Fig.~\ref{fig:arch}) is employed to close the gripper at a constant velocity of $v^g$, while the side trigger (in orange in Fig.~\ref{fig:arch}) is used to open the gripper similarly.  As soon as the back trigger is released, the gripper motion is stopped, thus allowing the human operator to modulate the gripper closure. Similar considerations apply for the side trigger.  
Note that the \textit{A} button's function to enable/disable the robot motion serves as a critical safety feature, preventing unintended movement by the operator. Furthermore, it grants the operators the flexibility to adjust their hand position should they approach the limits of their workspace. Once repositioned, they can resume providing velocity commands to the robot (see accompanying video).

\subsection{Partial autonomy for slippage prevention}
 
\begin{figure}
    \centering
    \includegraphics[width=0.9\linewidth]{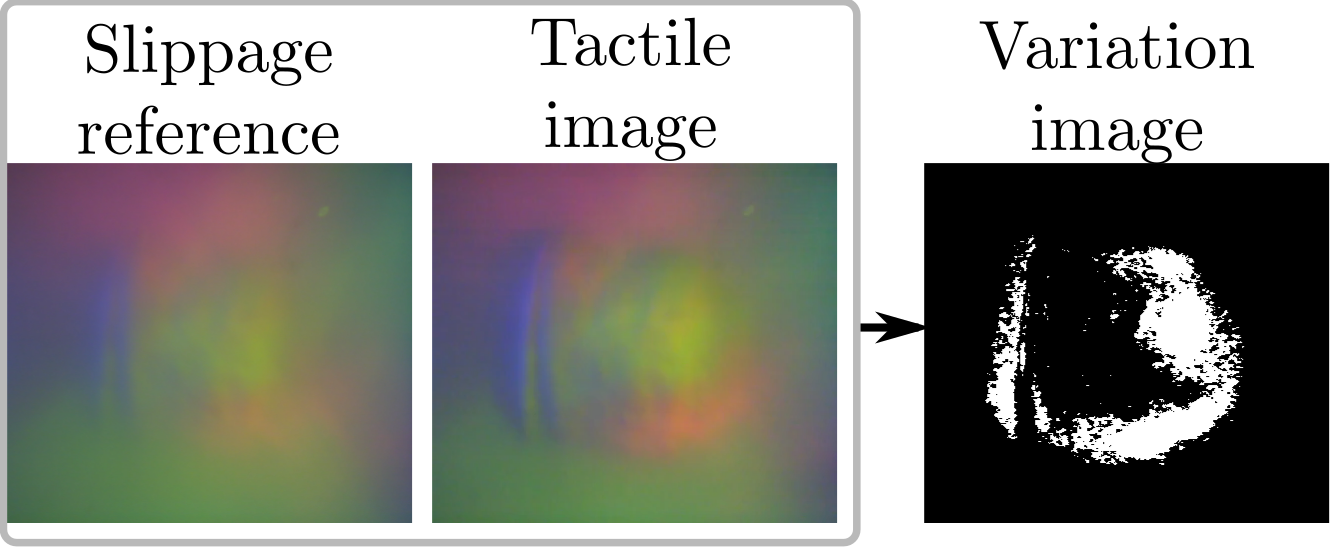}
    \caption{Example of the slippage reference image (left), tactile data (middle), and respective variation image (right) during a pistachio nut manipulation. 
    }
    \label{fig:slippage}
\end{figure}

The partial autonomy behavior assists the operator in preventing object slippage. Briefly,  in the event of automatic slip detection, the robot takes autonomous action to secure the grasp and prevent the object from slipping out of its grip. To realize the slip detection, we adopt a similar approach to the one described earlier for computing variation images. 
More specifically, when touch is detected on both sensors at time step $k$, i.e., $p^1_k>\epsilon_t$ and $p^2_k>\epsilon_t$, with $\epsilon_t$ a positive constant, the autonomous behavior, reported in Algorithm~\ref{alg:pa} is activated and an additional reference image $\bar{R}^s$  for each sensor $s$, referred to as slippage reference image, is computed by averaging $N$  consecutive tactile images (lines 3-4).  
Note that this reference image does not depict the background; instead, it represents the tactile data of the object currently being grasped as shown in Fig.~\ref{fig:slippage}. 
When the autonomous behavior is active, we calculate  at each time step~$k$ the variation ratios, denoted as $\bar{p}^s_k, \forall s$, with respect to the slippage reference images. This calculation is performed in accordance with Algorithm~\ref{alg:tcd} using the slippage reference images $\bar{R}^s$ (lines 6-7). 
A slippage event is detected if either $\bar{p}^1_k>\zeta^s$ or $\bar{p}^2_k>\zeta^s$ is observed, where $\zeta^s, $ $\forall s$,  is a time-varying threshold updated as follows (lines 5 and 14): 
$$\zeta^s=2^{{\# slip}}\epsilon_t,$$
with ${\#slip}$ representing the number of slippages detected since the autonomous behavior activation. 

If slippage is detected, the grasp is automatically tightened by a displacement $\delta_s$ (line 9).  After tightening, new slippage reference images are acquired (lines 10-11), and the thresholds $\zeta^s$ are increased (line 12). 
The reason for increasing the threshold is to dynamically adjust the sensitivity of slip detection over time~\cite{welle2023enabling}. As the object is grasped more tightly, a higher tolerance for slip detection is necessary. Without this adaptation, the system might continually detect slippage for minor changes, hindering the successful manipulation of the object. 
When the gripper is opened, i.e., the side trigger is pressed, the slippage prevention is deactivated. 

\begin{algorithm}
    \caption{Partial autonomy}\label{alg:pa}
    \begin{algorithmic}[1]
        \Require $T^s_k, \bar{R}^s, \bar{L}^s=[B^s_{k-c+1},..., B^s_{k-1}], \eta^s$ with $s=1,2$, $\epsilon_t,\delta_s,N, \text{initialize}$ 
        \If{initialize}
            \State $\bar{R}^1$ = compute reference image($N$)
            \State $\bar{R}^2$ = compute reference image($N$)
            \State $\#slip=0$
            \State $\zeta^1=\epsilon_t, \,\zeta^2=\epsilon_t$
        \EndIf
        \State $\bar{p}^1_k, \bar{B}^1_k=$tactile variation detection($T^1_k,\bar{L}^1,\bar{R}^1,\eta^1$)    
        \State $\bar{p}^2_k, \bar{B}^2_k=$tactile variation detection($T^2_k,\bar{L}^2,\bar{R}^2,\eta^2$)
        \If{$\bar{p}^1_k>\zeta^1$ or $\bar{p}^2_k>\zeta^2$}
            \State tighten grasp($\delta_s$)
            \State $\bar{R}^1$ = compute reference image($N$)
            \State $\bar{R}^2$ = compute reference image($N$)
            \State $\#slip=\#slip+1$
            \State $\zeta^s=2^{\#slip}\epsilon_t, \,\forall s$
        
        \EndIf  
    \end{algorithmic}
\end{algorithm}

\section{Experimental results}\label{sec:exp}
\begin{figure}
    \centering
    \includegraphics[width=0.9\linewidth]{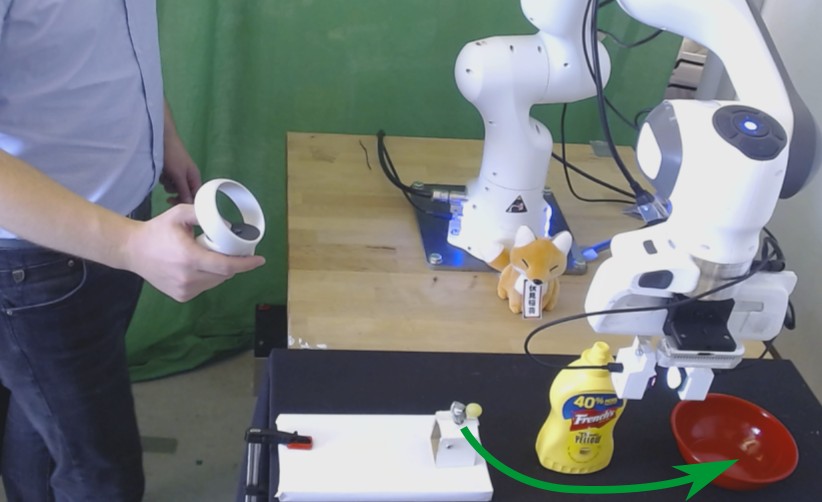}
    \caption{Experimental setup for T2H framework validation. The desired motion is highlighted with a green arrow. }
    \label{fig:exp}
\end{figure}

\begin{figure}
    \centering
    \includegraphics[width=1\linewidth]{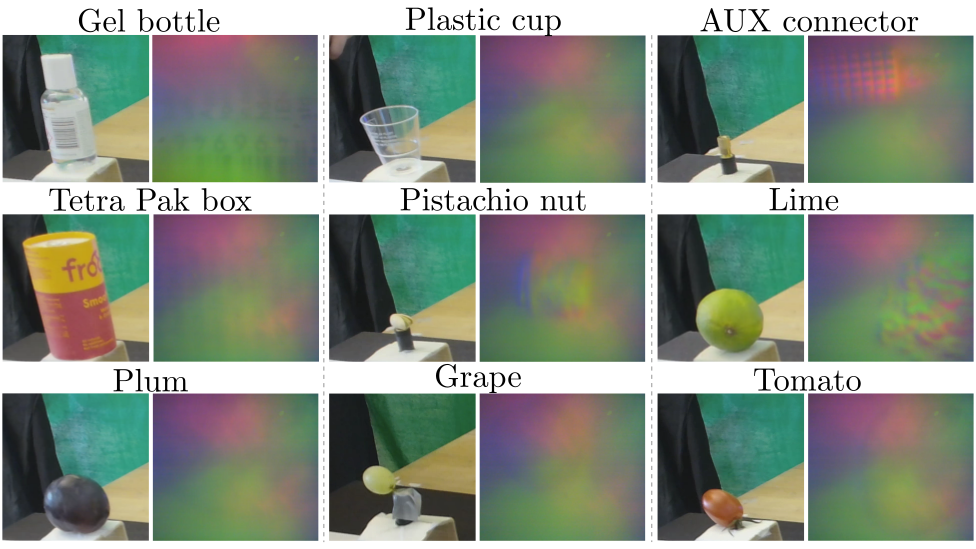}
    \vspace{-13pt}
    \caption{Set of objects along with the respective tactile data. }
    \label{fig:objects}
\end{figure}

The aim of the experiments is to teleoperate the robot in order to manipulate several objects, moving them from their initial positions to a  bowl, as illustrated in Fig.~\ref{fig:exp}. An additional obstacle object (a yellow mustard bottle) is introduced into the scene and must be navigated around. 
The objects under consideration consist of soft food items, that are lime, plum, grape, and tomato, as well as more rigid items, that are AUX connector, Tetra Pak box,  plastic gel bottle, plastic cup, and pistachio nut. The selected objects have distinct sizes, shapes, textures, and material properties, such as brittleness, presenting varied challenges for manipulation.  
The initial configurations of all objects are shown in  Fig.~\ref{fig:objects}. It is worth highlighting that in the initial configuration, the AUX connector is fully inserted in the respective port, requiring a certain force to be exerted for the initial unplugging phase. 
Similarly, the grapes and tomatoes are attached to the respective stems, requiring delicate manipulation to pick the fruit berry and detach it from the stem. While detaching, it is crucial to exert enough force to successfully separate the fruit, but it is equally important not to grasp too forcefully in order to prevent any damage to the delicate object. All the remaining objects do not require any unplugging/detaching operation before transportation. 
Three distinct operators participated in the experiments, one of whom was an experienced operator (specifically, an author of the paper) 
 while the others did not have familiarity with the T2H framwork. This enabled to assess the framework usability with non-experienced users. 
More in detail, the experienced operator performed the manipulation of each object three times without enabling the partial autonomy behavior and three times with it, while the non-experience user $1$ performed twice the manipulation of all objects without partial autonomy behavior and the non-experience user $2$ performed twice the manipulation of all objects with partial autonomy behavior. These resulted in a total of $90$ experiments. 
 We remark  that the reduced number of tests conducted by inexperienced users is attributed to the time-consuming nature of the tests. Nevertheless, their aim remains to validate the usability of the system for inexperienced users. Furthermore, note that in all the experiments the users are close to the robot performing the task and can visually monitor it. However, the T2H approach can also be adopted in remote teleoperation scenarios by providing the human operator with a video stream showing the robotic setup. 

All experimental videos, code, and setup instructions are available on the project website\footnote{\fontsize{6}{8}\label{foot:website}\url{https://vision-tactile-manip.github.io/teleop/}}.  Furthermore, the accompanying video reports examples of manipulation for all objects.


 \begin{table*}[]
 \centering
 \caption{Results obtained by the experienced (left) and non-experienced (right) users. Duration, minimum gripper opening, and $\#slip$ with (grey cells) and without (white cells) PA enabled are reported ($\#slip$ is only recorded with PA).   }
\vspace{-9pt}
 \setlength{\tabcolsep}{3.5pt}
 \resizebox{\textwidth}{!}{
\begin{tabular}{l|llllll||llllll|}
\cline{2-13}
                             & \multicolumn{6}{c||}{Experienced user}                                                                                                                                                                                                                                                                 & \multicolumn{6}{c|}{Non-experienced users}                                                                                                                                                                                                                                                              \\ \cline{2-13} 
                             & \multicolumn{2}{c|}{Duration {[}s{]}}                                                                                     & \multicolumn{2}{c|}{Min. opening {[}cm{]}}                                                               & \multicolumn{2}{c||}{$\#slip$}                                       & \multicolumn{2}{c|}{Duration {[}s{]}}                                                                                       & \multicolumn{2}{c|}{Min. opening {[}cm{]}}                                                               & \multicolumn{2}{c|}{$\#slip$}                                       \\ \hline
\multicolumn{1}{|l|}{Object} & \multicolumn{1}{c|}{\cellcolor[HTML]{EFEFEF}{\xmark} PA}         & \multicolumn{1}{c|}{\cellcolor[HTML]{FFFFFF}{\cmark} PA}        & \multicolumn{1}{c|}{\cellcolor[HTML]{EFEFEF}{\xmark} PA}          & \multicolumn{1}{c|}{{\cmark} PA}         & \multicolumn{1}{c|}{\cellcolor[HTML]{EFEFEF}{\xmark} PA} & \multicolumn{1}{c||}{{\cmark} PA}       & \multicolumn{1}{c|}{\cellcolor[HTML]{EFEFEF}{\xmark} PA}           & \multicolumn{1}{c|}{{\cmark} PA}        & \multicolumn{1}{c|}{\cellcolor[HTML]{EFEFEF}{\xmark} PA}          & \multicolumn{1}{c|}{{\cmark} PA}         & \multicolumn{1}{c|}{\cellcolor[HTML]{EFEFEF}{\xmark} PA} & \multicolumn{1}{c|}{{\cmark} PA}       \\ \hline
\multicolumn{1}{|l|}{Bottle} & \multicolumn{1}{l|}{\cellcolor[HTML]{EFEFEF}$29.4 \pm 2.1$} & \multicolumn{1}{l|}{\cellcolor[HTML]{FFFFFF}$28.1 \pm 5.2$} & \multicolumn{1}{l|}{\cellcolor[HTML]{EFEFEF}$3.16 \pm 0.52$} & \multicolumn{1}{l|}{$2.64 \pm 0.51$} & \multicolumn{1}{l|}{\cellcolor[HTML]{EFEFEF}$N.A.$} & $2.7 \pm 0.9$ & \multicolumn{1}{l|}{\cellcolor[HTML]{EFEFEF}$49.0 \pm 7.6$}   & \multicolumn{1}{l|}{\cellcolor[HTML]{FFFFFF}$43.3 \pm 4.3$} & \multicolumn{1}{l|}{\cellcolor[HTML]{EFEFEF}$3.24 \pm 0.33$} & \multicolumn{1}{l|}{$2.27 \pm 0.03$} & \multicolumn{1}{l|}{\cellcolor[HTML]{EFEFEF}$N.A.$} & $1.0 \pm 0.0$ \\ \hline
\multicolumn{1}{|l|}{Cup}    & \multicolumn{1}{l|}{\cellcolor[HTML]{EFEFEF}$31.4 \pm 3.1$} & \multicolumn{1}{l|}{\cellcolor[HTML]{FFFFFF}$30.8 \pm 3.6$} & \multicolumn{1}{l|}{\cellcolor[HTML]{EFEFEF}$3.96 \pm 0.09$} & \multicolumn{1}{l|}{$4.04 \pm 0.14$} & \multicolumn{1}{l|}{\cellcolor[HTML]{EFEFEF}$N.A.$} & $1.7 \pm 0.5$ & \multicolumn{1}{l|}{\cellcolor[HTML]{EFEFEF}$47.4 \pm 5.7$}   & \multicolumn{1}{l|}{\cellcolor[HTML]{FFFFFF}$36.4 \pm 0.5$} & \multicolumn{1}{l|}{\cellcolor[HTML]{EFEFEF}$3.75 \pm 0.13$} & \multicolumn{1}{l|}{$4.25 \pm 0.13$} & \multicolumn{1}{l|}{\cellcolor[HTML]{EFEFEF}$N.A.$} & $1.5 \pm 0.5$ \\ \hline
\multicolumn{1}{|l|}{AUX}    & \multicolumn{1}{l|}{\cellcolor[HTML]{EFEFEF}$41.3 \pm 7.6$} & \multicolumn{1}{l|}{\cellcolor[HTML]{FFFFFF}$36.9 \pm 3.5$} & \multicolumn{1}{l|}{\cellcolor[HTML]{EFEFEF}$0.90 \pm 0.05$} & \multicolumn{1}{l|}{$0.89 \pm 0.00$} & \multicolumn{1}{l|}{\cellcolor[HTML]{EFEFEF}$N.A.$} & $5.0 \pm 0.8$ & \multicolumn{1}{l|}{\cellcolor[HTML]{EFEFEF}$39.8 \pm 11.8$}  & \multicolumn{1}{l|}{\cellcolor[HTML]{FFFFFF}$44.0 \pm 9.3$} & \multicolumn{1}{l|}{\cellcolor[HTML]{EFEFEF}$0.72 \pm 0.03$} & \multicolumn{1}{l|}{$0.83 \pm 0.05$} & \multicolumn{1}{l|}{\cellcolor[HTML]{EFEFEF}$N.A.$} & $6.5 \pm 1.5$ \\ \hline
\multicolumn{1}{|l|}{Box}    & \multicolumn{1}{l|}{\cellcolor[HTML]{EFEFEF}$33.0 \pm 2.5$} & \multicolumn{1}{l|}{\cellcolor[HTML]{FFFFFF}$26.6 \pm 1.4$} & \multicolumn{1}{l|}{\cellcolor[HTML]{EFEFEF}$5.04 \pm 0.34$} & \multicolumn{1}{l|}{$5.19 \pm 0.06$} & \multicolumn{1}{l|}{\cellcolor[HTML]{EFEFEF}$N.A.$} & $1.0 \pm 0.0$ & \multicolumn{1}{l|}{\cellcolor[HTML]{EFEFEF}$28.8 \pm 5.9$}   & \multicolumn{1}{l|}{\cellcolor[HTML]{FFFFFF}$40.5 \pm 0.4$} & \multicolumn{1}{l|}{\cellcolor[HTML]{EFEFEF}$5.09 \pm 0.09$} & \multicolumn{1}{l|}{$4.61 \pm 0.58$} & \multicolumn{1}{l|}{\cellcolor[HTML]{EFEFEF}$N.A.$} & $1.0 \pm 0.0$ \\ \hline
\multicolumn{1}{|l|}{Nut}    & \multicolumn{1}{l|}{\cellcolor[HTML]{EFEFEF}$35.4 \pm 1.8$} & \multicolumn{1}{l|}{\cellcolor[HTML]{FFFFFF}$26.1 \pm 2.3$} & \multicolumn{1}{l|}{\cellcolor[HTML]{EFEFEF}$1.53 \pm 0.22$} & \multicolumn{1}{l|}{$1.53 \pm 0.34$} & \multicolumn{1}{l|}{\cellcolor[HTML]{EFEFEF}$N.A.$} & $2.7 \pm 1.2$ & \multicolumn{1}{l|}{\cellcolor[HTML]{EFEFEF}$52.7 \pm 13.2$}  & \multicolumn{1}{l|}{\cellcolor[HTML]{FFFFFF}$50.7 \pm 3.1$} & \multicolumn{1}{l|}{\cellcolor[HTML]{EFEFEF}$1.07 \pm 0.17$} & \multicolumn{1}{l|}{$1.44 \pm 0.20$} & \multicolumn{1}{l|}{\cellcolor[HTML]{EFEFEF}$N.A.$} & $2.5 \pm 0.5$ \\ \hline
\multicolumn{1}{|l|}{Lime}   & \multicolumn{1}{l|}{\cellcolor[HTML]{EFEFEF}$28.4 \pm 3.0$} & \multicolumn{1}{l|}{\cellcolor[HTML]{FFFFFF}$25.5 \pm 3.0$} & \multicolumn{1}{l|}{\cellcolor[HTML]{EFEFEF}$4.95 \pm 0.09$} & \multicolumn{1}{l|}{$4.83 \pm 0.14$} & \multicolumn{1}{l|}{\cellcolor[HTML]{EFEFEF}$N.A.$} & $1.0 \pm 0.0$ & \multicolumn{1}{l|}{\cellcolor[HTML]{EFEFEF}$38.3 \pm 0.1$}   & \multicolumn{1}{l|}{\cellcolor[HTML]{FFFFFF}$45.1 \pm 0.4$} & \multicolumn{1}{l|}{\cellcolor[HTML]{EFEFEF}$4.44 \pm 0.02$} & \multicolumn{1}{l|}{$4.75 \pm 0.00$} & \multicolumn{1}{l|}{\cellcolor[HTML]{EFEFEF}$N.A.$} & $1.0 \pm 0.0$ \\ \hline
\multicolumn{1}{|l|}{Plum}   & \multicolumn{1}{l|}{\cellcolor[HTML]{EFEFEF}$27.6 \pm 2.3$} & \multicolumn{1}{l|}{\cellcolor[HTML]{FFFFFF}$27.7 \pm 4.3$} & \multicolumn{1}{l|}{\cellcolor[HTML]{EFEFEF}$4.59 \pm 0.20$} & \multicolumn{1}{l|}{$4.33 \pm 0.19$} & \multicolumn{1}{l|}{\cellcolor[HTML]{EFEFEF}$N.A.$} & $1.0 \pm 0.0$ & \multicolumn{1}{l|}{\cellcolor[HTML]{EFEFEF}$43.4 \pm 1.0$}   & \multicolumn{1}{l|}{\cellcolor[HTML]{FFFFFF}$37.7 \pm 0.8$} & \multicolumn{1}{l|}{\cellcolor[HTML]{EFEFEF}$3.78 \pm 0.08$} & \multicolumn{1}{l|}{$3.98 \pm 0.02$} & \multicolumn{1}{l|}{\cellcolor[HTML]{EFEFEF}$N.A.$} & $1.0 \pm 0.0$ \\ \hline
\multicolumn{1}{|l|}{Grape}  & \multicolumn{1}{l|}{\cellcolor[HTML]{EFEFEF}$44.3 \pm 3.4$} & \multicolumn{1}{l|}{\cellcolor[HTML]{FFFFFF}$38.0 \pm 3.2$} & \multicolumn{1}{l|}{\cellcolor[HTML]{EFEFEF}$1.68 \pm 0.11$} & \multicolumn{1}{l|}{$1.75 \pm 0.18$} & \multicolumn{1}{l|}{\cellcolor[HTML]{EFEFEF}$N.A.$} & $1.0 \pm 0.0$ & \multicolumn{1}{l|}{\cellcolor[HTML]{EFEFEF}$102.5 \pm 22.1$} & \multicolumn{1}{l|}{\cellcolor[HTML]{FFFFFF}$33.7 \pm 2.0$} & \multicolumn{1}{l|}{\cellcolor[HTML]{EFEFEF}$1.30 \pm 0.00$} & \multicolumn{1}{l|}{$1.56 \pm 0.09$} & \multicolumn{1}{l|}{\cellcolor[HTML]{EFEFEF}$N.A.$} & $1.0 \pm 0.0$ \\ \hline
\multicolumn{1}{|l|}{Tomato} & \multicolumn{1}{l|}{\cellcolor[HTML]{EFEFEF}$33.8 \pm 5.2$} & \multicolumn{1}{l|}{\cellcolor[HTML]{FFFFFF}$31.8 \pm 1.0$} & \multicolumn{1}{l|}{\cellcolor[HTML]{EFEFEF}$2.73 \pm 0.28$} & \multicolumn{1}{l|}{$3.20 \pm 0.10$} & \multicolumn{1}{l|}{\cellcolor[HTML]{EFEFEF}$N.A.$} & $1.0 \pm 0.0$ & \multicolumn{1}{l|}{\cellcolor[HTML]{EFEFEF}$60.0 \pm 29.3$}   & \multicolumn{1}{l|}{\cellcolor[HTML]{FFFFFF}$35.1 \pm 1.6$} & \multicolumn{1}{l|}{\cellcolor[HTML]{EFEFEF}$2.48 \pm 0.65$} & \multicolumn{1}{l|}{$2.56 \pm 0.15$} & \multicolumn{1}{l|}{\cellcolor[HTML]{EFEFEF}$N.A.$} & $1.0 \pm 0.0$ \\ \hline
\end{tabular}
}
\label{tab:exp}
\end{table*}
\noindent
\textbf{Implementation details:}
The parameters of the T2H framework  for the experiments were selected as follows. For the mapping to teleoperation controller and partial autonomy, we used gain $\alpha = 1000$, generated the reference images using a frame count of $N=10$, tuned the automatic noise threshold over $K=100$ frames, set the touch threshold to $\epsilon_t=0.01$, and considered $c=2$ consecutive frames for variation detection.
For the mapping to robot inputs, the maximum linear velocity was set to $v^l_{max}=0.1$ m/s, and the maximum rotational velocity to $1$ rad/s, while we disregarded linear velocities below $0.005$ m/s and rotational velocities below $0.05$ rad/s.
Furthermore, we used $w=5$ for the velocity averaging filter. 
The closing gripper speed was set to \mbox{$v_g=0.005$} m/s, while the displacement in the event of slippage was $\delta_s=0.001$~m.
 
\noindent \textbf{Results: }
As demonstrated in the experiment videos available on the project website, all  three operators were able to successfully complete all teleoperation tasks with various objects, including delicate ones, without causing any damage.  
Table~\ref{tab:exp} summarizes the results obtained on the manipulation of all objects with  (white cells) and without (grey cells) partial autonomy (PA) for slippage prevention enabled. Specifically, we report the duration, the minimum gripper opening, and the number of slippages (when PA is enabled only) achieved by the experienced (left) and non-experienced (right) operators. Average and standard deviation values are shown. 
 Among the considered metrics, 
the minimum gripper opening serves as an indicator of object compression, with a larger opening implying less compression applied. Note that this metric is an absolute value and is not normalized relative to the (unknown) object width. Therefore, it must be compared on a per-object basis. 
 For the experienced user, we can observe that the AUX connector and the grape berry require the longest manipulation times due to the challenging initial unplugging/detaching operations and the delicate nature of the grape.
Across all objects, the autonomous slippage prevention behavior consistently contributes to the reduction of the time to complete the manipulation task. 
This is particularly evident for the AUX connector, pistachio nut, Tetra Pak box, and grape berry,
with the AUX connector exhibiting the highest frequency of slippage prevention activation (i.e., showing the highest value of $\#slip$). This is motivated by the requirement of applying a certain amount of force to unplug the connector, and insufficient force can lead to slippage. In such instances, the autonomous slippage detection and grasp tightening mechanisms significantly enhance the efficiency of task completion.
In addition, we can observe that the PA behavior also leads to a reduction in object compression. This is evident in the fact that, on average, the minimum gripper opening required to complete the task is either lower or comparable. 
As far as the non-experienced users are concerned, similar trends to those observed for the experienced user are obtained, only resulting in average longer completion times and higher object compression. These results highlight the usability of the T2H framework as well as the benefits conferred by the slippage prevention autonomous behavior.

\begin{figure}
    \centering
    \includegraphics[width=1\linewidth]{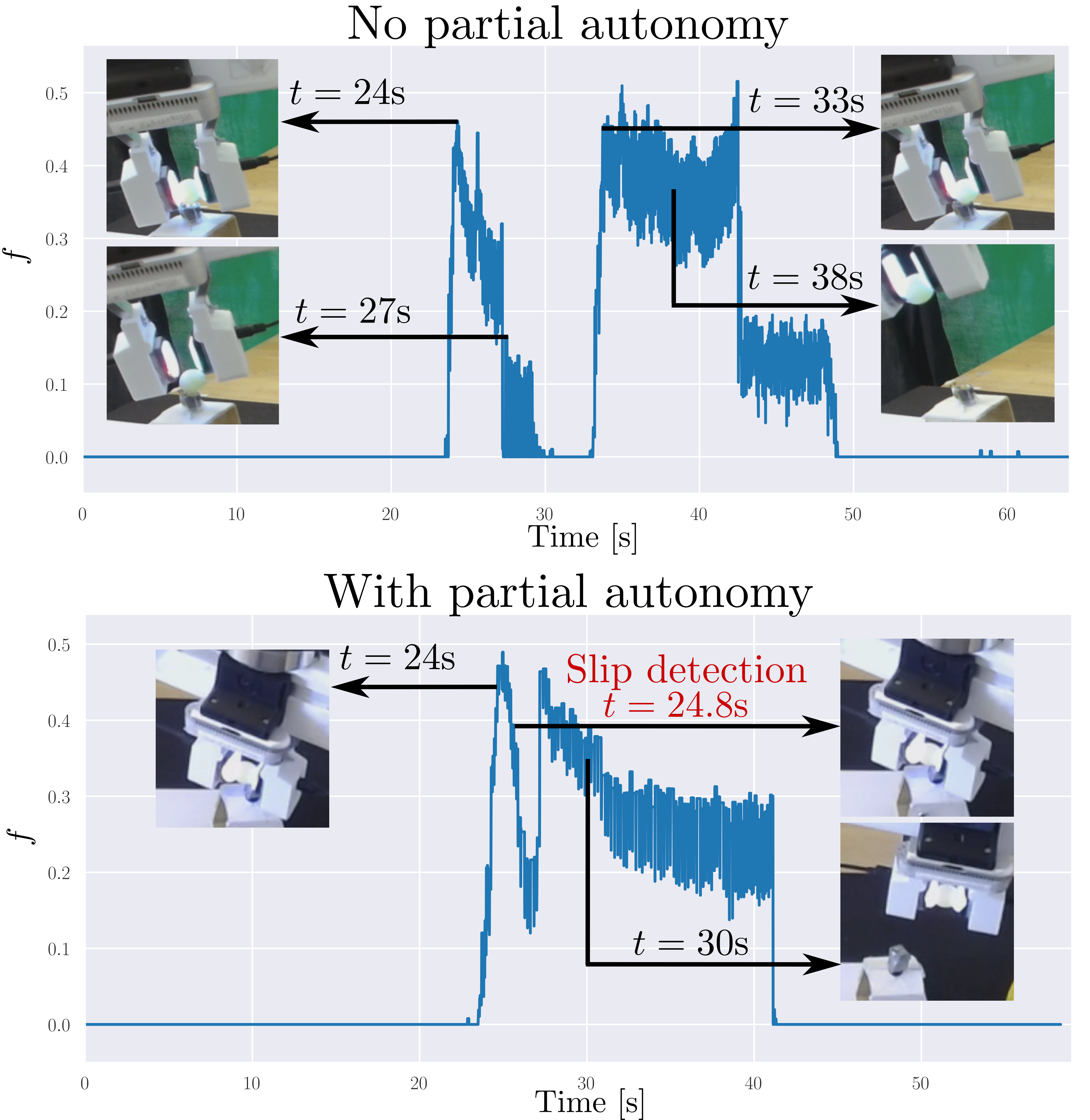}
    \vspace{-9pt}
    \caption{Haptic feedback obtained during the grape manipulation by the expert user without (top) and with (bottom) partial autonomy enabled. 
    }
    \label{fig:feedback}
\end{figure}

Figure~\ref{fig:feedback} reports two illustrative examples of the haptic feedback $f$ obtained by the experienced operator during the manipulation of the grape berry,  both with (bottom) and without (top) enabling the partial autonomy behavior. Relevant screenshots depicting the manipulation process are included. 
In particular, in the absence of partial autonomy, the operator initiates the grasping of the berry at approximately $t=24$~s, and a respective increase in haptic feedback amplitude is observed. 
However, as an attempt is made to move upwards, the berry slips from the grasp (at approximately $t=27$~s), as indicated by the drop in the feedback signal.
Subsequently, the operator makes a second attempt to secure the berry (at about $t=33$~s), ultimately completing the manipulation (as shown in the screenshot at $t=38$~s). When the partial autonomy is active,  the operator performs the grasping of the berry at about $t=24$~s, resulting in a corresponding increase in haptic feedback.
However, in this case,  automatic slippage detection occurs at approximately $t=24.8$~s, triggering an autonomous tightening of the grasp. This intervention effectively prevents any slippage of the grape berry and enables the operator to complete the manipulation task (as shown in the screenshot at $t=30$~s)  without having to re-grasp the grape berry. Further examples  are available in the accompanying video and on the project website. 

\section{Conclusion}
In this study, we proposed the novel T2H framework, providing haptic feedback based on vision-based tactile sensors for teleoperation.
 We realized T2H with low-cost consumer-grade hardware and made the source code public.
The framework leverages pixel-wise variation in tactile images compared to background reference images to provide vibration feedback, enhancing operator perception and control during object manipulation.  
Additionally, a partial autonomy feature is designed to prevent slippage during manipulation tasks.
The system's effectiveness was demonstrated across a range of objects with diverse physical properties, including fragile items, without causing damage. Usability was confirmed across operators with varying experience levels. Future work aims to introduce proactive behaviors by predicting human intentions, integrate a GUI with the headset   and conduct extensive user studies involving multiple operators.

\bibliographystyle{ieeetr}
\balance
\bibliography{references}

\begin{thebibliography}{10}

\bibitem{yamaguchi2019recent}
A.~Yamaguchi and C.~G. Atkeson, ``Recent progress in tactile sensing and sensors for robotic manipulation: can we turn tactile sensing into vision?,'' {\em Advanced Robotics}, vol.~33, no.~14, pp.~661--673, 2019.

\bibitem{vbsensors_review}
S.~Zhang, Z.~Chen, Y.~Gao, W.~Wan, J.~Shan, H.~Xue, F.~Sun, Y.~Yang, and B.~Fang, ``Hardware technology of vision-based tactile sensor: A review,'' {\em IEEE Sensors Journal}, vol.~22, no.~22, pp.~21410--21427, 2022.

\bibitem{yamaguchi2019tactile}
A.~Yamaguchi and C.~G. Atkeson, ``Tactile behaviors with the vision-based tactile sensor fingervision,'' {\em International Journal of Humanoid Robotics}, vol.~16, no.~03, p.~1940002, 2019.

\bibitem{zhu_icra2022}
X.~Zhu, S.~Jain, M.~Tomizuka, and J.~Van~Baar, ``Learning to synthesize volumetric meshes from vision-based tactile imprints,'' in {\em International Conference on Robotics and Automation (ICRA)}, pp.~4833--4839, 2022.

\bibitem{welle2023enabling}
M.~C. Welle, M.~Lippi, H.~Lu, J.~Lundell, A.~Gasparri, and D.~Kragic, ``Enabling robot manipulation of soft and rigid objects with vision-based tactile sensors,'' in {\em International Conference on Automation Science and Engineering (CASE)}, 2023.

\bibitem{teleop_survey_tro}
K.~Darvish, L.~Penco, J.~Ramos, R.~Cisneros, J.~Pratt, E.~Yoshida, S.~Ivaldi, and D.~Pucci, ``Teleoperation of humanoid robots: A survey,'' {\em IEEE Transactions on Robotics}, 2023.

\bibitem{el2020review}
I.~El~Rassi and J.-M. El~Rassi, ``A review of haptic feedback in tele-operated robotic surgery,'' {\em Journal of medical engineering \& technology}, vol.~44, no.~5, pp.~247--254, 2020.

\bibitem{bolopion_TASE2013}
A.~Bolopion and S.~Régnier, ``A review of haptic feedback teleoperation systems for micromanipulation and microassembly,'' {\em IEEE Transactions on Automation Science and Engineering}, vol.~10, no.~3, pp.~496--502, 2013.

\bibitem{lambeta2020digit}
M.~Lambeta, P.-W. Chou, S.~Tian, B.~Yang, B.~Maloon, V.~R. Most, D.~Stroud, R.~Santos, A.~Byagowi, G.~Kammerer, {\em et~al.}, ``Digit: A novel design for a low-cost compact high-resolution tactile sensor with application to in-hand manipulation,'' {\em IEEE Robotics and Automation Letters}, vol.~5, no.~3, pp.~3838--3845, 2020.

\bibitem{darvish2023teleoperation}
K.~Darvish, L.~Penco, J.~Ramos, R.~Cisneros, J.~Pratt, E.~Yoshida, S.~Ivaldi, and D.~Pucci, ``Teleoperation of humanoid robots: A survey,'' {\em IEEE Transactions on Robotics}, 2023.

\bibitem{xie2023design}
Y.~Xie, X.~Hou, and S.~Wang, ``Design of a novel haptic joystick for the teleoperation of continuum-mechanism-based medical robots,'' {\em Robotics}, vol.~12, no.~2, p.~52, 2023.

\bibitem{ishiguro2020bilateral}
Y.~Ishiguro, T.~Makabe, Y.~Nagamatsu, Y.~Kojio, K.~Kojima, F.~Sugai, Y.~Kakiuchi, K.~Okada, and M.~Inaba, ``Bilateral humanoid teleoperation system using whole-body exoskeleton cockpit tablis,'' {\em IEEE Robotics and Automation Letters}, vol.~5, no.~4, pp.~6419--6426, 2020.

\bibitem{gliesche2020kinesthetic}
P.~Gliesche, T.~Krick, M.~Pfingsthorn, S.~Drolshagen, C.~Kowalski, and A.~Hein, ``Kinesthetic device vs. keyboard/mouse: a comparison in home care telemanipulation,'' {\em Frontiers in Robotics and AI}, vol.~7, p.~561015, 2020.

\bibitem{nandikolla2022teleoperation}
V.~Nandikolla and D.~A. Medina~Portilla, ``Teleoperation robot control of a hybrid eeg-based bci arm manipulator using ros,'' {\em Journal of Robotics}, vol.~2022, 2022.

\bibitem{wozniak2023virtual}
M.~Wozniak, C.~T. Chang, M.~B. Luebbers, B.~Ikeda, M.~Walker, E.~Rosen, and T.~R. Groechel, ``Virtual, augmented, and mixed reality for human-robot interaction (vam-hri),'' in {\em Companion of the 2023 ACM/IEEE International Conference on Human-Robot Interaction}, pp.~938--940, 2023.

\bibitem{barentine2021vr}
C.~Barentine, A.~McNay, R.~Pfaffenbichler, A.~Smith, E.~Rosen, and E.~Phillips, ``A vr teleoperation suite with manipulation assist,'' in {\em Companion of the 2021 ACM/IEEE International Conference on Human-Robot Interaction}, pp.~442--446, 2021.

\bibitem{xu2022shared}
S.~Xu, S.~Moore, and A.~Cosgun, ``Shared-control robotic manipulation in virtual reality,'' {\em arXiv preprint arXiv:2205.10564}, 2022.

\bibitem{moletta2023virtual}
M.~Moletta, M.~K. Wozniak, M.~C. Welle, and D.~Kragic, ``A virtual reality framework for human-robot collaboration in cloth folding,'' {\em arXiv preprint arXiv:2305.07493}, 2023.

\bibitem{chandan2021arroch}
K.~Chandan, V.~Kudalkar, X.~Li, and S.~Zhang, ``Arroch: Augmented reality for robots collaborating with a human,'' in {\em 2021 IEEE International Conference on Robotics and Automation (ICRA)}, pp.~3787--3793, IEEE, 2021.

\bibitem{wozniak2023happily}
M.~K. Wozniak, R.~Stower, P.~Jensfelt, and A.~Pereira, ``Happily error after: Framework development and user study for correcting robot perception errors in virtual reality,'' {\em arXiv preprint arXiv:2306.14589}, 2023.

\bibitem{ostanin2018interactive}
M.~Ostanin and A.~Klimchik, ``Interactive robot programing using mixed reality,'' {\em IFAC-PapersOnLine}, vol.~51, no.~22, pp.~50--55, 2018.

\bibitem{lorenzini2022performance}
M.~Lorenzini, S.~Ciotti, J.~M. Gandarias, S.~Fani, M.~Bianchi, and A.~Ajoudani, ``Performance analysis of vibrotactile and slide-and-squeeze haptic feedback devices for limbs postural adjustment,'' in {\em 2022 31st IEEE International Conference on Robot and Human Interactive Communication (RO-MAN)}, pp.~707--713, IEEE, 2022.

\bibitem{chua2023modular}
Z.~Chua and A.~M. Okamura, ``A modular 3-degrees-of-freedom force sensor for robot-assisted minimally invasive surgery research,'' {\em Sensors}, vol.~23, no.~11, p.~5230, 2023.

\bibitem{smith2022feeling}
A.~Smith, B.~Ward-Cherrier, A.~Etoundi, and M.~J. Pearson, ``Feeling the pressure: The influence of vibrotactile patterns on feedback perception.,'' in {\em 2022 IEEE/RSJ International Conference on Intelligent Robots and Systems (IROS)}, pp.~634--640, IEEE, 2022.

\bibitem{luzfeeling}
R.~Luz, A.~Pereira, J.~Corujeira, T.~Krueger, J.~Beck, E.~den Exter, T.~Chupin, J.~L. Silva, and R.~Ventura, ``Feeling the slope? teleoperation of a mobile robot using a 7dof haptic device with attitude feedback,''

\bibitem{pocius2020communicating}
R.~Pocius, N.~Zamani, H.~Culbertson, and S.~Nikolaidis, ``Communicating robot goals via haptic feedback in manipulation tasks,'' in {\em Companion of the 2020 ACM/IEEE International Conference on Human-Robot Interaction}, pp.~591--593, 2020.

\bibitem{al2023resolving}
Z.~Al-Saadi, Y.~M. Hamad, Y.~Aydin, A.~Kucukyilmaz, and C.~Basdogan, ``Resolving conflicts during human-robot co-manipulation,'' in {\em Proceedings of the 2023 ACM/IEEE International Conference on Human-Robot Interaction}, pp.~243--251, 2023.

\bibitem{hinchet2018dextres}
R.~Hinchet, V.~Vechev, H.~Shea, and O.~Hilliges, ``Dextres: Wearable haptic feedback for grasping in vr via a thin form-factor electrostatic brake,'' in {\em Proceedings of the 31st Annual ACM Symposium on User Interface Software and Technology}, pp.~901--912, 2018.

\bibitem{palmer2022haptic}
J.~E. Palmer, M.~Sarac, A.~A. Garza, and A.~M. Okamura, ``Haptic feedback relocation from the fingertips to the wrist for two-finger manipulation in virtual reality,'' in {\em 2022 IEEE/RSJ International Conference on Intelligent Robots and Systems (IROS)}, pp.~628--633, IEEE, 2022.

\bibitem{coffey2022collaborative}
M.~Coffey and A.~Pierson, ``Collaborative teleoperation with haptic feedback for collision-free navigation of ground robots,'' in {\em 2022 IEEE/RSJ International Conference on Intelligent Robots and Systems (IROS)}, pp.~8141--8148, IEEE, 2022.

\bibitem{stanley2012evaluation}
A.~A. Stanley and K.~J. Kuchenbecker, ``Evaluation of tactile feedback methods for wrist rotation guidance,'' {\em IEEE Transactions on Haptics}, vol.~5, no.~3, pp.~240--251, 2012.

\bibitem{andrussow2023minsight}
I.~Andrussow, H.~Sun, K.~J. Kuchenbecker, and G.~Martius, ``Minsight: A fingertip-sized vision-based tactile sensor for robotic manipulation,'' {\em Advanced Intelligent Systems}, p.~2300042, 2023.

\bibitem{yuan2017gelsight}
W.~Yuan, S.~Dong, and E.~H. Adelson, ``Gelsight: High-resolution robot tactile sensors for estimating geometry and force,'' {\em Sensors}, vol.~17, no.~12, p.~2762, 2017.

\bibitem{cao2023vis2hap}
G.~Cao, J.~Jiang, N.~Mao, D.~Bollegala, M.~Li, and S.~Luo, ``Vis2hap: Vision-based haptic rendering by cross-modal generation,'' {\em arXiv preprint arXiv:2301.06826}, 2023.

\end{thebibliography}

\end{document}